\documentclass{article}
\usepackage{spconf,amsmath,graphicx,booktabs,multirow,tabularx,bm,float}

\title{ Using Auxiliary tasks in multimodal fusion of Wav2vec 2.0 and BERT for Multimodal Emotion Recognition}

\name{Dekai Sun, Yancheng He, Jiqing Han\sthanks{Corresponding authors: jqhan@hit.edu.cn}}

\address{Harbin Institute of Technology, Harbin, China}

\begin{document}
\maketitle 
%————————————————————————————————%
\begin{abstract}
The lack of data and the difficulty of multimodal fusion have always been challenges for multimodal emotion recognition (MER). 
% In this paper, we propose to use pretrained models as upstream network, and design auxiliary tasks for downstream network while modal fusion is performed. First, we leverage pretrained models, wav2vec 2.0 for audio modality and BERT for text modality. Second, the feature information of different modalities is integrated through a multi-layer cross-attention mechanism. Furthermore, we design two auxiliary tasks to effectively fuse the two modalities.
In this paper, we propose to use pretrained models as upstream network, wav2vec 2.0 for audio modality and BERT for text modality, and finetune them in downstream task of MER to cope with the lack of data. For the difficulty of multimodal fusion, we use a K-layer multi-head attention mechanism as a downstream fusion module. Starting from the MER task itself, we design two auxiliary tasks to alleviate the insufficient fusion between modalities and guide the network to capture and align emotion-related features.
Compared to the previous state-of-the-art models, we achieve a better performance by 78.42\% Weighted Accuracy (WA) and 79.71\% Unweighted Accuracy (UA) on the IEMOCAP dataset.
\end{abstract}
%————————————————————————————————%
%————————————————————————————————%
\section{Introduction}
\label{sec:intro}
Multimodal emotion recognition is a significant capability in human-machine interaction and has attracted widespread attention in industry and academia. As we all know that emotions are expressed in extremely complex and ambiguous ways, perhaps through linguistic content, speech intonation, facial expression and body actions. There have been many related studies on text emotion recognition \cite{adoma2020comparative, acheampong2021transformer}, and also on audio emotion recognition \cite{shen2011automatic, kishore2013emotion, siriwardhana2020jointly}. However, by observing these results, the research on single modality has reached a certain bottleneck which leads to increasing attention devoted to the use of multimodal approach. Some studies propose that the information of different modalities is often complementary and verified, and the full use of the information of different modalities can help the model to better learn the key content  \cite{soleymani2011multimodal, mittal2020m3er}.

%----------------------
In recent years, pretrained self-supervised learning has performed prominently in several research fields such as natural language processing (NLP) \cite{devlin2018bert} and automatic speech recognition (ASR) \cite{baevski2020wav2vec}. For the multimodal emotion recognition (MER) task, there are also studies that have done a lot of exploration on the basis of pretrained models. 
For the first time, Siriwardhana et al. \cite{siriwardhana2020jointly} jointly finetuned modality-specific “BERT-like” pretrained Self Supervised Learning (SSL) architectures to represent both audio and text modalities for the task of MER. Similarly, Yang et al. \cite{yang22q_interspeech} also proposed to finetune two pretrained self-supervised learning models (Text-RoBERTa and Speech-RoBERTa) for MER. Based on pretrained models, Zhao et al. \cite{zhao22k_interspeech} explored Multi-level fusion approaches, including coattention-based early fusion and late fusion with the models trained on both embeddings. Compared with the MCSAN \cite{sun2021multimodal} using traditional features (MFCC \& GloVe) for modal fusion, the works mentioned above have greatly improved performance. From the perspective of making full use of contextual data, Wu et al. \cite{wu2021emotion} took advantage of contextual information and proposed a two-branch neural network structure including time synchronous branch and time asynchronous branch. By modifying the structure of network, SMCN \cite{hou2022multi} realize multi-modal alignment which can capture the global connections without interfering with unimodal learning. 
% Different from previous work, we design auxiliary tasks to help the deep fusion of modalities and guide the network to learn features and alignment information related to emotion itself.
However, these previous works focused more on sophisticated fusion structure design and the use of larger and stronger pretrained models, or the use of contextual information that breaks data constraints.
They did not start from the MER task itself to explore the bottleneck of insufficient fusion, or capture the feature of emotion itself and the alignment of emotion in different modalities.
We believe that the parameters of the network are already sufficient, and the complex fusion module design has not brought enough benefits. Thus, we hope to guide the model to fully exploit the potential of the fusion module by designing just the right auxiliary tasks.
%----------------------

\begin{figure}[t]
\centering
\includegraphics[width=0.35\textwidth]{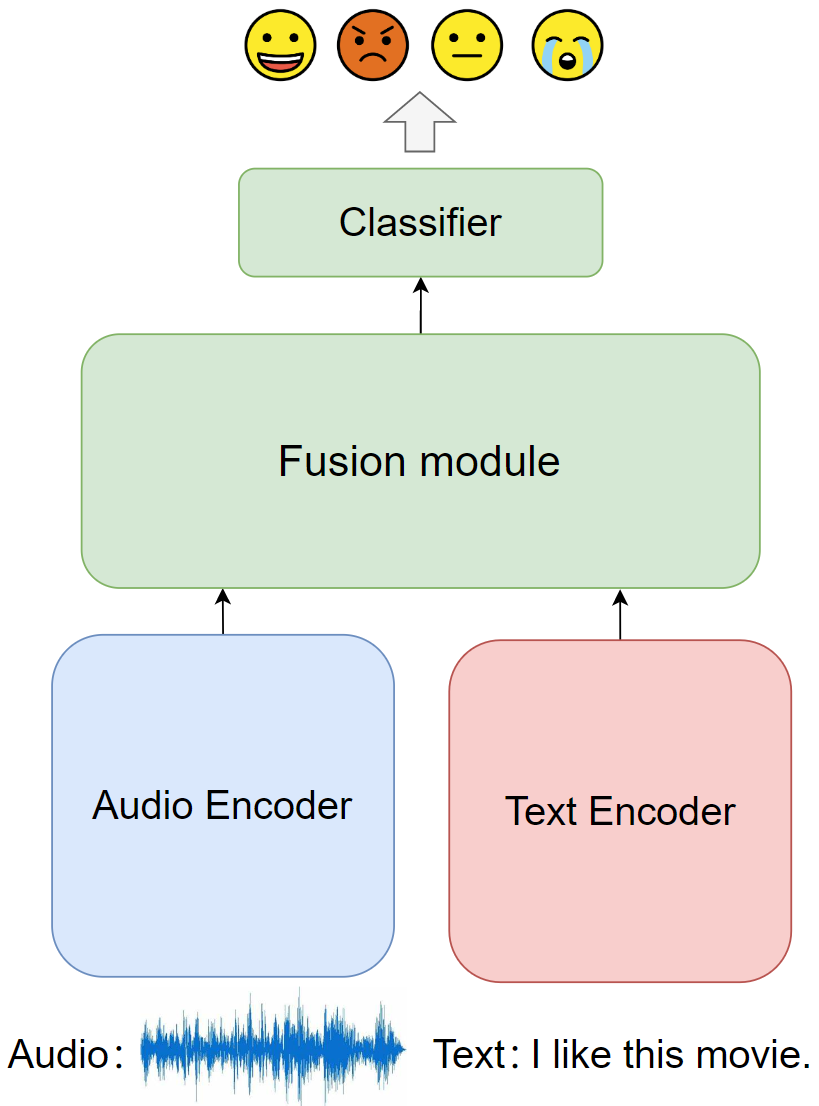}
\caption{Our framework}
\label{fig:figure-1}
\end{figure}

In this work, we propose a modular end-to-end approach for the MER task. The general framework is shown in figure \ref{fig:figure-1}. First, we learn the semantic information of the respective modalities through the pretrained models, wav2vec 2.0 \cite{baevski2020wav2vec} for audio modality and BERT \cite{devlin2018bert} for text modality. Then, we map text and audio modal feature information into a unified semantic vector space through a k-layer cross-attention mechanism for more adequate modal fusion. 
Furthermore, we design two auxiliary tasks to help fully fuse the features of the two modalities and learn the alignment information of the emotion itself between different modalities. 
In the first one, we randomly recombine text and audio modalities and let the model to predict the combination of the two modalities through the vector obtained by fusion. This decoupling of multimodal data enables the model to see more complex input combinations, and the constraint of this auxiliary task forces the network to not ignore the role of any modality in the task of MER.
In the second one, we randomly replace one of the modalities with other data of the same emotion category, and hope that the model can capture the feature  related to emotion and the alignment information beyond the content itself. 

We comprehensively evaluated the performance of the model proposed on the IEMOCAP dataset in terms of average weighted accuracy (WA) and unweighted accuracy (UA). In additional, we compared it with the SOTA methods and presented relevant ablation experiments that illustrate the effectiveness of each module.
%————————————————————————————————%
%————————————————————————————————%
\section{Method}
\label{sec:format}
The framework of our proposed model is showned in Figure \ref{fig:figure-1}, which consists of three modules, i.e., text encoder, audio encoder, and fusion module. 
%We will describe these three modules in detail.

\subsection{Text Encoder}
The emergence of BERT has brought NLP into a new era, and gradually refreshed the effect of multiple NLP domain tasks. And “Pretrain + Finetune” has gradually become a new paradigm. Pre-training models such as BERT can be used to transform text into word vectors with contextual semantic information. In this paper, we choose bert-base-uncased\footnote{https://huggingface.com/bert-base-uncased} as the text modal encoder, which consists 12 layers of transformer encoder. It converts the text into 768-dimensional vectors, which are fed into the fusion module. During training, we also finetune its weights to make it more suitable for our multimodal emotion recognition task.

\subsection{Audio Encoder}
We choose wav2vec2-base\footnote{https://huggingface.co/facebook/wav2vec2-base} as the audio modality encoder, which consists of feature encoder, contextualized representations with Transformers, and quantization module. The base model contains 12 transformer blocks, and it is pretrained in Librispeech corpus containing 960 hours of 16kHz speech. It is able to learn 768-dimensional latent representation directly from raw audio every 20ms (16Khz sampling rate). We also finetune its parameters during training similar to BERT. 

\subsection{Fusion Module}
The fusion module is based on the multi-head cross attention mechanism \cite{vaswani2017attention}. In addition, two auxiliary tasks (Section \ref{sec:Auxiliary Tasks}) help the model to better handle the feature relationship between the two modalities. Figure \ref{fig:figure-3} shows the specific details of the fusion module, and each layer of the fusion module consists of two branches, which have the same structure but different Q, K, and V. In addition, we use residual linking to reduce the loss of information of the original modalities. The calculation process of multi-head cross attention is as follows:
\begin{equation}
  F_a = F_a + Attention_{at}(Q_a,K_t,V_t)
  \label{eq3}
\end{equation}
\begin{equation}
   Attention_{at}(Q_a,K_t,V_t) = softmax(\dfrac{Q_aK_t^T}{\sqrt{d_{K_t}}})V_t
  \label{eq4}
\end{equation}
\begin{equation}
  F_t = F_t + Attention_{ta}(Q_t,K_a,V_a)
  \label{eq1}
\end{equation}
\begin{equation}
  Attention_{ta}(Q_t,K_a,V_a) = softmax(\dfrac{Q_tK_a^T}{\sqrt{d_{K_a}}})V_a
  \label{eq2}
\end{equation}
where subscript $a$ represents audio modality and subscript $t$ represents text modality. $d_{K_a}$ and $d_{K_t}$ represent dimension of the embeddings. $F_t$ : $(B, T_t, C)$ is the text feature outputed by BERT,  and $F_a$ : $(B, T_a, C)$ is the audio feature outputed by Wav2vec 2.0. $Q_a$, $K_a$, $V_a$ are given here (same of $Q_t$, $K_t$, $V_t$):
\begin{equation}
   Q_a = W_QF_a + b_a^Q
  \label{eq5}
\end{equation}
\begin{equation}
   K_a = W_KF_a + b_a^K
  \label{eq6}
\end{equation}
\begin{equation}
   V_a = W_VF_a + b_a^V
  \label{eq7}
\end{equation}
Finally, we average pooling $F_a$ and $F_t$ in the time dimension, and concatenate them in the feature dimension to obtain the fusion embedding $(B, 2C)$, which is sent to the classifier to get the emotion category.

\subsection{Auxiliary Tasks}
\label{sec:Auxiliary Tasks}
In order to help the model fully fuse the features of the two modalities and learn the alignment information of the emotion itself between different modalities, we design two auxiliary modal interaction tasks.

\subsubsection{Auxiliary Task1}
\begin{figure}[t]
\centering
\includegraphics[width=0.48\textwidth]{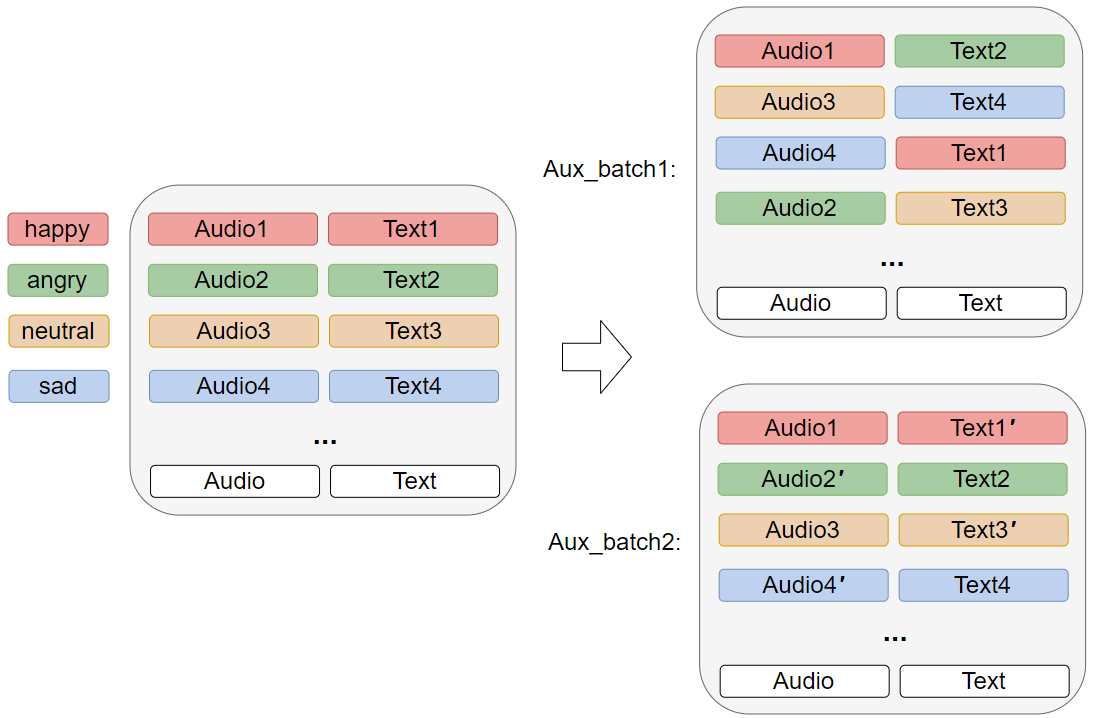}
\caption{Auxiliary Batch}
\label{fig:figure-2}
\end{figure}

\begin{figure}[t]
\centering
\includegraphics[width=0.48\textwidth]{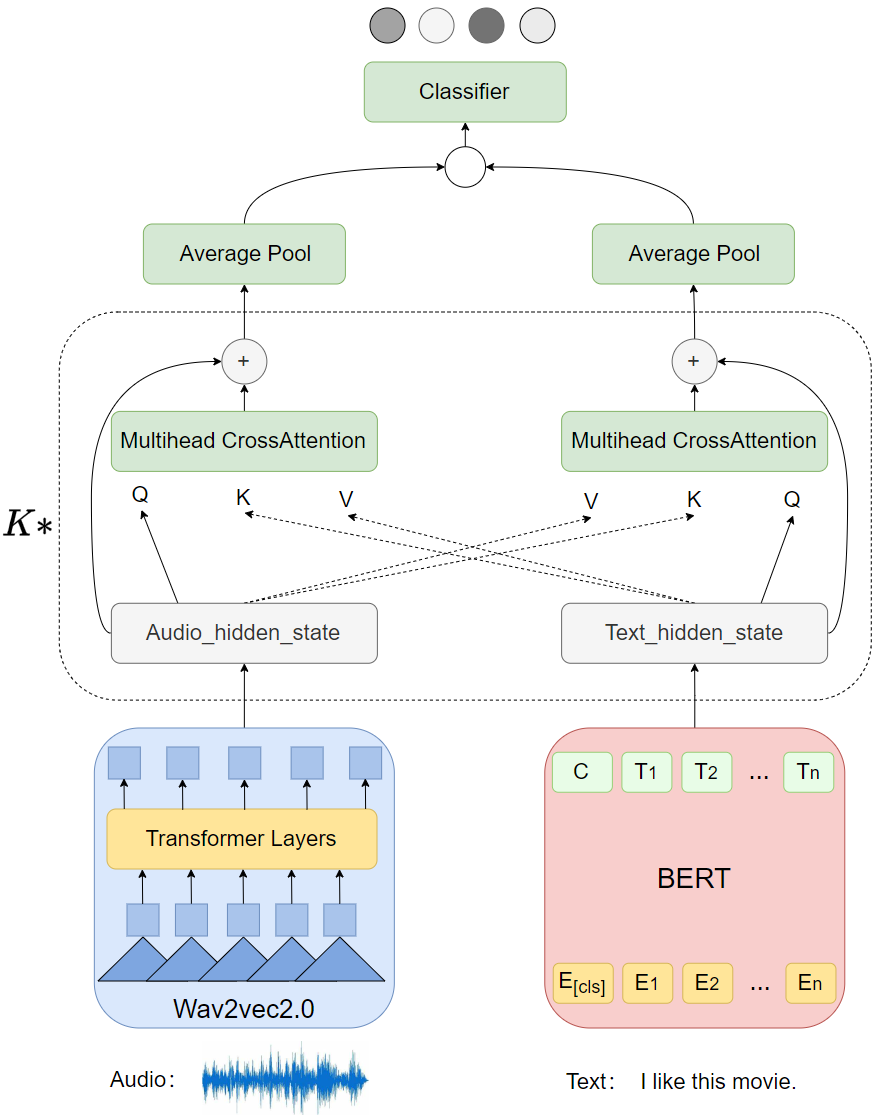}
\caption{Proposed model structure}
\label{fig:figure-3}
\end{figure}

In MER tasks, audio and text have the same semantics. In the modal fusion of the downstream network, we analyze that the reason for insufficient fusion comes from the fact that the overall emotional orientation can be obtained just from the information of one modality. In some cases, this approach leads to the right results. But for complex cases, we want the network to be more ``humble", making full use of the information of the two modalities.
%audio (semantic external emotional features), text (semantic internal emotional features).
% In the previous experiment, we found that when the information of two modalities is ambiguous, the model can not correctly predict the overall emotional category presented. We analyzed that this is because the training data are all text and audio with the same semantic information (that is, the text is converted from audio), which is strongly related in semantic level. The model can correctly predict the overall emotional category from part of the information of one modality. Therefore, it is easy to ignore other characteristic information. 
As shown in Figure \ref{fig:figure-2}, we decouple the pairs of \{Audio, Text\} in a batch of data, and then randomly scramble and recombine them to get Aux\_batch1. During the training process, we not only let the model predict the emotion category of the original data pair, but also predict the combined category of this reorganized data pair \{Audio, Text\} (a total of $emotion\_num \times emotion\_num$ kinds), and its label ($label_{new}$) is defined as follows:
% So in the auxiliary task 1, as shown in Figure \ref{fig:figure-2}, we decoupled each pair of datas, then reorder and combine text and audio data pairs in the same batch. In the training process, we not only let the model predict the emotional category of the original data pairs, but also predict the emotional category of the text and audio modules of such reorganized data pairs (that is, add a 16 category task). The formula is as follows.
\begin{equation}
   label_{original} = label_a = label_t
  \label{eq8}
\end{equation}
\begin{equation}
   label_{new} = label_a \times emotion\_nums + label_t
  \label{eq9}
\end{equation}
The main task MER requires the downstream network to receive the features from the two modalities and output the emotion category, while the auxiliary task 1 requires the downstream network to predict not only the emotion but also the combination of the two modalities according to the fusion embedding. It forces the downstream network to not ignore any modal information during the feature fusion process of the two modalities, that is, both modal information contributes to the final fusion embedding.
% In this way, the model is forced to see more feature information combination to enhance the robustness of the model, and at the same time, the characteristics of the two modes can be learned more fully, and the experimental results also verified our idea.
%————————————————————————————————%
%————————————————————————————————%
\subsubsection{Auxiliary Task2}
In order to guide the fusion network to learn the alignment information of emotion itself between different modalities, we break the strong semantic correlation between modalities. As shown in Figure \ref{fig:figure-2}, for the pairs of \{Audio, Text\} in a batch of data, we randomly replace one of the modalities (Audio or Text) with other data of the same emotion category. 
% To solve the problem that the strong correlation between text and voice leads to insufficient learning of each part of the modality, we designed the auxiliary task 2. In this task, we randomly replace one of the modalities (text and audio pair) with other data of the same emotion category. We hope to break the strong content correlation between the two modality information. As shown in Figure \ref{fig:figure-2}, In a batch, randomly replace one of the modal information of text and audio to meet the requirement that each data pair is modal information with the same emotion but different content.
In Aux\_batch2, different modalities have same emotional label but different semantics. We hope that the fusion network can focus on the features of emotion itself in different modalities and align them. At the same time, the model can better learn common features of the same emotion category.
% In this way, the model can better capture the features related to the emotion category recognition task. For example, the model can learn that the high intonation is related to the word "happy" in the text, rather than just aligning the content.(待斟酌)
% At the same time, the model can better learn the common features of the same kind of emotion category. For example, for the emotion category of happy, the text may generally contain the word "happy".

%————————————————————————————————%

%————————————————————————————————%
\section{Experimental setup}
\label{sec:format}

\subsection{Dataset}
The dataset used in the experiment is the Interactive Emotional Dyadic Motion Cap-ture (IEMOCAP) \cite{busso2008iemocap}, which is a dialogues dataset and performs improvised and scripts by 10 actors. The 10 actors are divided into 5 sessions, and every session consists of 1 male and 1 female. There are a total of 7529 utterances in \textbf{IEMOCAP} (happy 595, excited 1,041, angry 1,103, sad 1,084, neutral 1,708, frustration 1,849, fear 40, surprise 107, disgust 2). To be consistent and compare with previous studies \cite{pepino21_interspeech}, only utterances with ground truth labels belonging to “angry”, “happy”, “excited”, “sad”, and “neutral” were used. The “excited” class was merged with “happy” to better balance the size of each emotion class, which results in a total of 5,531 utterances (happy 1,636, angry 1,103, sad 1,084, neutral 1,708).
\subsection{Implementation Details}
In order to fully evaluate our proposed model and maintain the same test conditions as previous studies \cite{wu2021emotion}, a leave-one-session-out 5-fold cross-validation (CV) configuration was implemented to evaluate our model. We divide IEMOCAP into five folds according to sessions in our experiments. At each fold we keep one session for testing, and other sessions are used for training. Therefore, for each fold we can get one result, and we take the average of the results as the final result of our experiments.

We implement our model within the PyTorch framework and select the AdamW \cite{loshchilov2017decoupled} optimizer for model optimization with a learning rate of $1 \times 10^{-5}$, where cross attention had 8 heads.
 
\section{Results}
\label{sec:typestyle}
%————————————————————————————————%
Table \ref{tab:example} shows the performance of our method on audio-only, text-only, and  multimodal (audio and text) emotion recognition tasks. Compared with a single modality, we simply concatenate the features of the two modalities and feed them into a downstream network constructed with a fully connected layer (FC), which improves the performance by about 6\%. Further, we use the single-layer (K=1) multi-head cross-attention downstream network in Figure 3 for modality fusion, which achieves WA : 77.19\%, UA : 78.47\%. In the current state, we also verify the gains of Auxiliary Task 1 and Auxiliary Task 2, of which Auxiliary Task 2 has the better performance. We also try to use both auxiliary tasks with performance WA : 78.34\%, UA : 79.59\%.
\begin{table}[t]
  \caption{Weighted Accuracy (WA) and Unweighted Accuracy (UA) of the 5-fold CV results using single modality and multi modality.(FC - Fully Connected; CA - Multi-Head Cross Attention (K=1); Aux1 - Auxiliary Task1; Aux2 - Auxiliary Task2.)}
  \label{tab:example}
  \centering
  \begin{tabular}{ lcc }
    \toprule
    \textbf{Methods} & \textbf{WA($\%$)} & \textbf{UA($\%$)} \\
    \midrule
    \textbf{Text-only} \\
    BERT            & $70.53$  & $71.79$             \\
    \midrule
    \textbf{Audio-only} \\
    Wav2vec2        & $69.92$  & $70.68$       \\
    \midrule
    \textbf{Audio and Text} \\
    BERT+Wav2vec2+FC     & $76.24$  & $77.20$              \\
    BERT+Wav2vec2+CA    & $77.19$  & $78.47$              \\
    BERT+Wav2vec2+CA+Aux1   & $77.67$  & $79.16$              \\
    BERT+Wav2vec2+CA+Aux2   & $78.11$  & $79.47$              \\
    \textbf{BERT+Wav2vec2+CA+Aux1\&2}  & \bm{$78.34$}  & \bm{$79.59$}   \\
    \bottomrule
  \end{tabular}
\end{table}
%————————————————————————————————%

Table \ref{tab:example2} shows that when both auxiliary tasks are used simultaneously, the effect of multi-head cross-attention layer K on the performance of emotion recognition task. When K is 2, we get the best performance WA : 78.42\%, UA : 79.71\%. We found that with the introduction of auxiliary tasks, the overall training objective of the model became difficult to achieve. By appropriately increasing the number of layers in the downstream network, we could obtain better performance. However, due to the limited size of the IEMOCAP dataset, continuously increasing the number of network layers will make it difficult to fully train the network parameters, resulting in performance degradation. The performance of previous state-of-the-art multimodal models is mentioned in Table \ref{tab:example3}, and our proposed method has better performance than previous works.
\begin{table}[t]
  \caption{Performance with different K (the number of layers of Multi-Head Cross Attention (CA)).}
  \label{tab:example2}
  \centering
  \begin{tabular}{ lccc }
    \toprule
    \textbf{Methods} & \textbf{K} & \textbf{WA($\%$)} & \textbf{UA($\%$)} \\
    BERT+Wav2vec2+CA+Aux1\&2   &$1$  & $78.34$  & $79.59$              \\
    \textbf{BERT+Wav2vec2+CA+Aux1\&2}   &\bm{$2$}  & \bm{$78.42$} & \bm{$79.71$}              \\
    BERT+Wav2vec2+CA+Aux1\&2   &$3$  & $77.68$  & $79.41$              \\
    \bottomrule
  \end{tabular}
\end{table}
%————————————————————————————————%

\begin{table}[H]
  \caption{Comparison of the 5-fold CV results of previous state-of-the-art multimodal models and our model on the IEMOCAP.}
  \label{tab:example3}
  \centering
  \begin{tabular}{ lcc }
  \toprule
    \textbf{Methods} & \textbf{WA($\%$)} & \textbf{UA($\%$)} \\
    \midrule
    BERT + Wav2vec2 \cite{zhao22k_interspeech}              & $-$  & $76.31$              \\
    RoBERTa-text\&audio \cite{yang22q_interspeech}              & $77.70$  & $78.50$              \\
    BERT + FBK \cite{wu2021emotion}              & $77.57$  & $78.41$              \\
    SMCN \cite{hou2022multi}              & $75.60$  & $77.60$              \\
    BERT + FBK \cite{morais2022speech}              & $70.56$  & $71.46$              \\
    MCSAN \cite{sun2021multimodal}              & $61.20$  & $56.00$              \\
    \textbf{Our proposed (best)}         & \bm{$78.42$}  & \bm{$79.71$} \\
    \bottomrule
  \end{tabular}
\end{table}
%————————————————————————————————%

\section{Conclusion}
\label{sec:majhead}
% In this paper, we proposed to use wav2vec 2.0 and BERT as upstream network for multimodal emotion recognition tasks, and a K-layer downstream network based on multi-head cross-attention mechanism for modalities fusion, in addition, we also explored the use of auxiliary tasks for better fusion of modalities.
In this paper, we propose to use wav2vec 2.0 and BERT as upstream network and K-layer downstream network based on multi-head cross-attention mechanism for multimodal emotion recognition task. In addition, we design two auxiliary tasks for the model to help the audio and text be fully integrated, and capture and align the features of emotion itself in different modalities. Finally our method outperforms the previous work on the 5-fold CV result of IEMOCAP, achieved the state-of-the-art, WA : 78.42\%, UA : 79.71\%.
% In the future, we intend to \textit{(1)} explore other auxiliary tasks related to emotion to help the model to not only learn features strongly related to emotion and better deep fusion of multiple modal features and \textit{(2)} support multimodal emotion recognition in dialogue.

\bibliographystyle{IEEEbib}
\bibliography{strings,refs}
%————————————————————————————————%
\end{document}